\documentclass[10pt,twocolumn,letterpaper]{article}

\usepackage[pagenumbers]{cvpr} 

\usepackage{graphicx}
\usepackage{amsmath}
\usepackage{amssymb}
\usepackage{booktabs}
\usepackage{multirow}
\usepackage{multicol}
\usepackage{tabularx}
\usepackage{amsmath}
\usepackage{amssymb}
\usepackage{algorithm}
\usepackage{algpseudocode}
\usepackage[pagebackref,breaklinks,colorlinks,allcolors=cvprblue]{hyperref}
\usepackage{bm}
\usepackage{soul}
\usepackage{makecell}
\usepackage[capitalize]{cleveref}

\definecolor{cnumber}{RGB}{204, 204, 204}

\usepackage{pifont}
\usepackage{circledtext}
\usepackage{pgffor}

\crefname{section}{Sec.}{Secs.}
\Crefname{section}{Section}{Sections}
\Crefname{table}{Table}{Tables}
\crefname{table}{Tab.}{Tabs.}
\newcommand{\Input}{\State \textbf{Input:} }
\newcommand{\Output}{\State \textbf{Output:} }
\newcommand{\abbr}{AdaCM$^2$}

\usepackage{tikz}
\newcommand*\Circled[1]{
	\tikz[baseline=(char.base)]{\node[
        shape=circle, draw=none,  thick, 
        fill=gray!40,inner sep=0.9pt] (char) 
    {\textcolor{black}{#1}}; 
}}

\definecolor{cvprblue}{rgb}{0.21,0.49,0.74}
\usepackage[pagebackref,breaklinks,colorlinks,allcolors=cvprblue]{hyperref}

\def\paperID{***} 
\def\confName{CVPR}
\def\confYear{2025}

\title{\abbr: On Understanding Extremely Long-Term Video with \underline{Ada}ptive \underline{C}ross-\underline{M}odality \underline{M}emory Reduction}

\author{Yuanbin Man$^{\text{1}}$, Ying Huang$^{\text{1}}$, Chengming Zhang$^{\text{2}}$, Bingzhe Li$^{\text{3}}$, Wei Niu$^{\text{4}}$, Miao Yin$^{\text{1}{\dagger}}$\\
{$^\text{1}$Department of CSE, University of Texas at Arlington, $^\text{2}$Department of CS, University of Houston,}\\
{$^\text{3}$Department of CS, University of Texas at Dallas, $^\text{4}$School of Computing, University of Georgia}\\
{\tt\small \texttt{\{yuanbin.man, ying.huang\}}@uta.edu, czhang48@uh.edu,} \\
{\tt\small  bingzhe.li@utdallas.edu, wniu@uga.edu, miao.yin@uta.edu}
}

\begin{document}
\maketitle
\begin{abstract}
\let\thefootnote\relax\footnotetext{\hspace{-5mm}$^\dagger$Corresponding author.}The advancements in large language models (LLMs) have propelled the improvement of video understanding tasks by incorporating LLMs with visual models. However, most existing LLM-based models (e.g., VideoLLaMA, VideoChat) are constrained to processing short-duration videos. Recent attempts to understand long-term videos by extracting and compressing visual features into a fixed memory size. Nevertheless, those methods leverage only visual modality to merge video tokens and overlook the correlation between visual and textual queries, leading to difficulties in effectively handling complex question-answering tasks. To address the challenges of long videos and complex prompts, we propose \abbr, which, for the first time, introduces an adaptive cross-modality memory reduction approach to video-text alignment in an auto-regressive manner on video streams. Our extensive experiments on various video understanding tasks, such as video captioning, video question answering, and video classification, demonstrate that \abbr~achieves state-of-the-art performance across multiple datasets while significantly reducing memory usage. Notably, it achieves a 4.5\% improvement across multiple tasks in the LVU dataset with a GPU memory consumption reduction of up to 65\%.
\end{abstract}    
\section{Introduction}
\label{sec:intro}

Video understanding is an important task in computer vision and artificial intelligence, which involves processing and reasoning over visual and textual information. While the recent success of large language models (LLMs) ~\cite{ouyang2022instructgpt, radford2019gpt2,brown2020gpt3,touvron2023llama} has significantly improved video-language models~\cite{miech2019howto100m, lu2019vilbert}, prior work has primarily focused on short video understanding tasks, typically with videos ranging from 5 to 15 seconds. However, long-term video understanding~\cite{lvu2021}, a sub-technique that develops models to process richer information, has played a crucial role in real-world applications such as movie analysis and video retrieval. Unfortunately, it poses significant challenges as video length increases, especially the \textit{large memory consumption} challenge. The number of frames the model must process grows rapidly, leading to substantial memory consumption, thereby preventing prior approaches from processing such long videos.

\begin{figure}[!t]
    \centering
    \includegraphics[width=\linewidth]{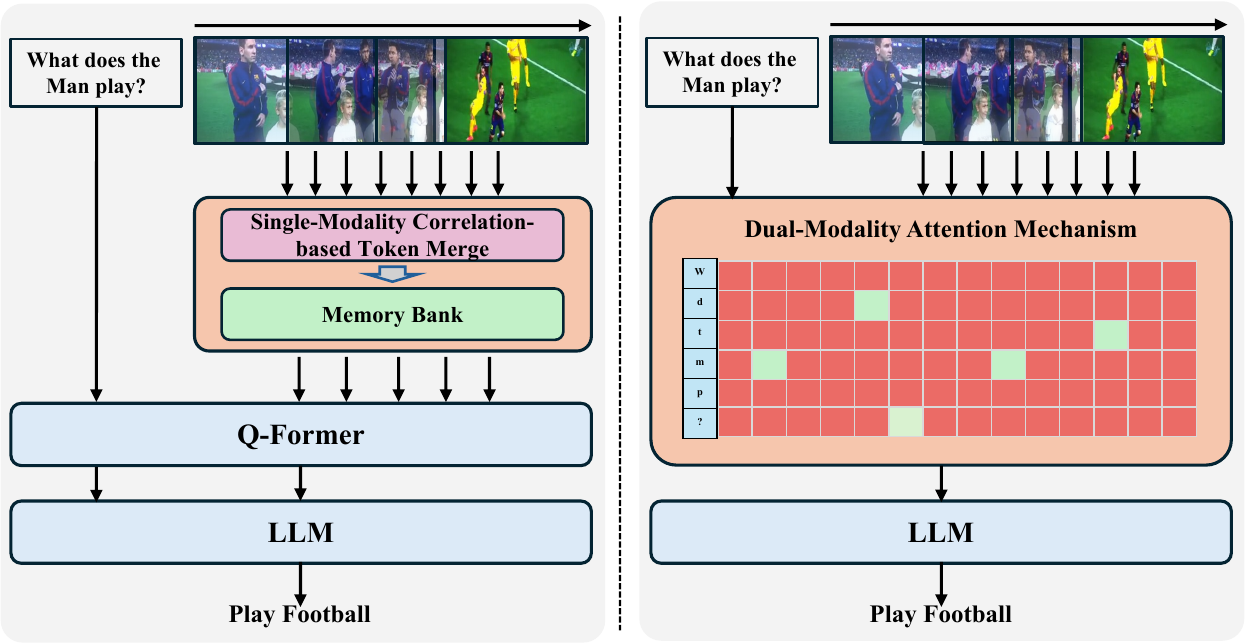}
    \vspace{-5mm}
    \caption{(Left) Existing approaches compress visual features of videos via single-modality correlation; (Right) Our \textbf{\abbr} reduces video memory adaptively based on cross-modality attention.}
    \label{fig:overview}
\end{figure}

\begin{figure*}[!t]
    \centering
    \includegraphics[width=0.9\linewidth]{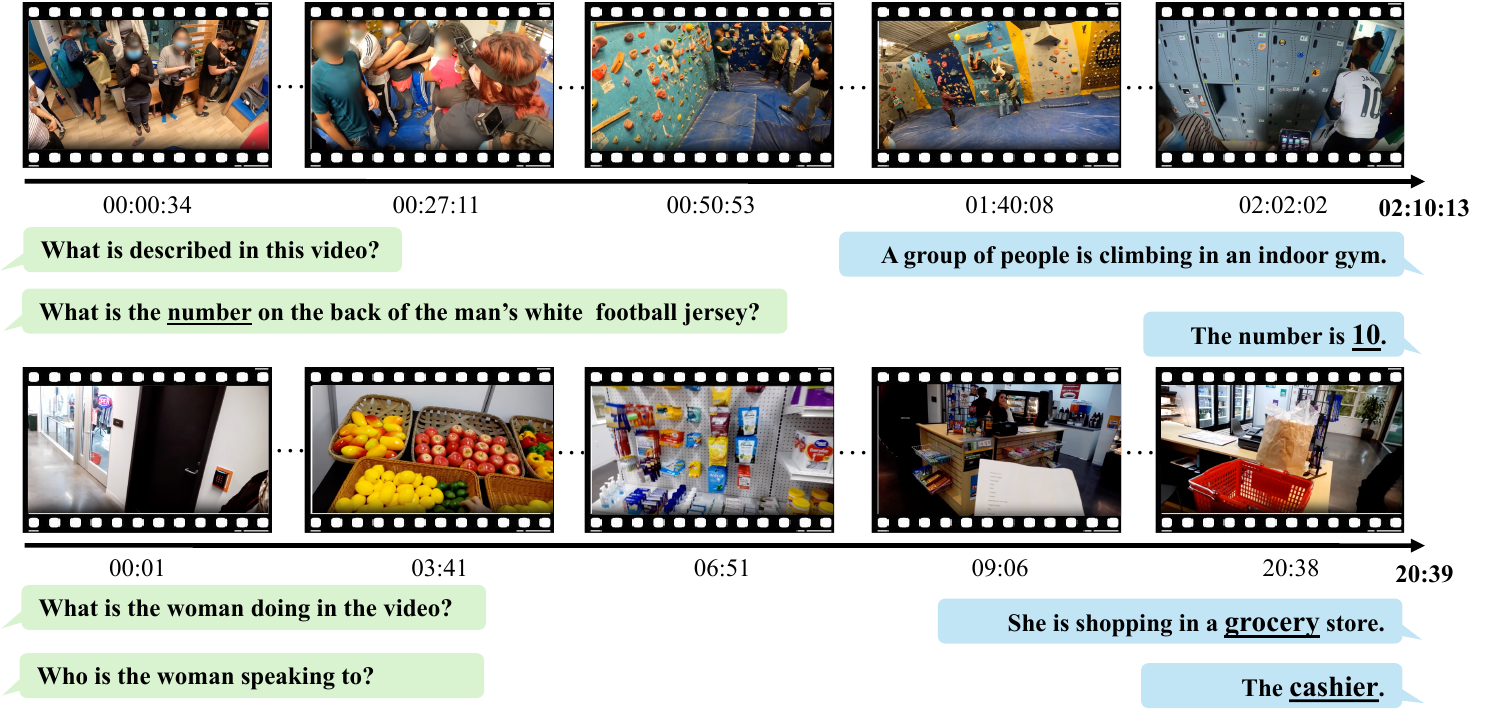} 
    \label{fig:your_label}
    \vspace{-1mm}
    \caption{The case study of \abbr~\textbf{zero-shot} on Ego4D\cite{grauman2022ego4d} dataset. As shown, \abbr~can 1) summarize an extremely long video lasting \textbf{over 2 hours} with limited memory consumption and identify the number on a person’s back at the end accurately, 2) answer questions related to a mid-length video spanning \textbf{more than 20 minutes}.}
\end{figure*}

\begin{figure*}[!ht]  
    \begin{subfigure}{0.48\textwidth}  
        \includegraphics[width=\linewidth]{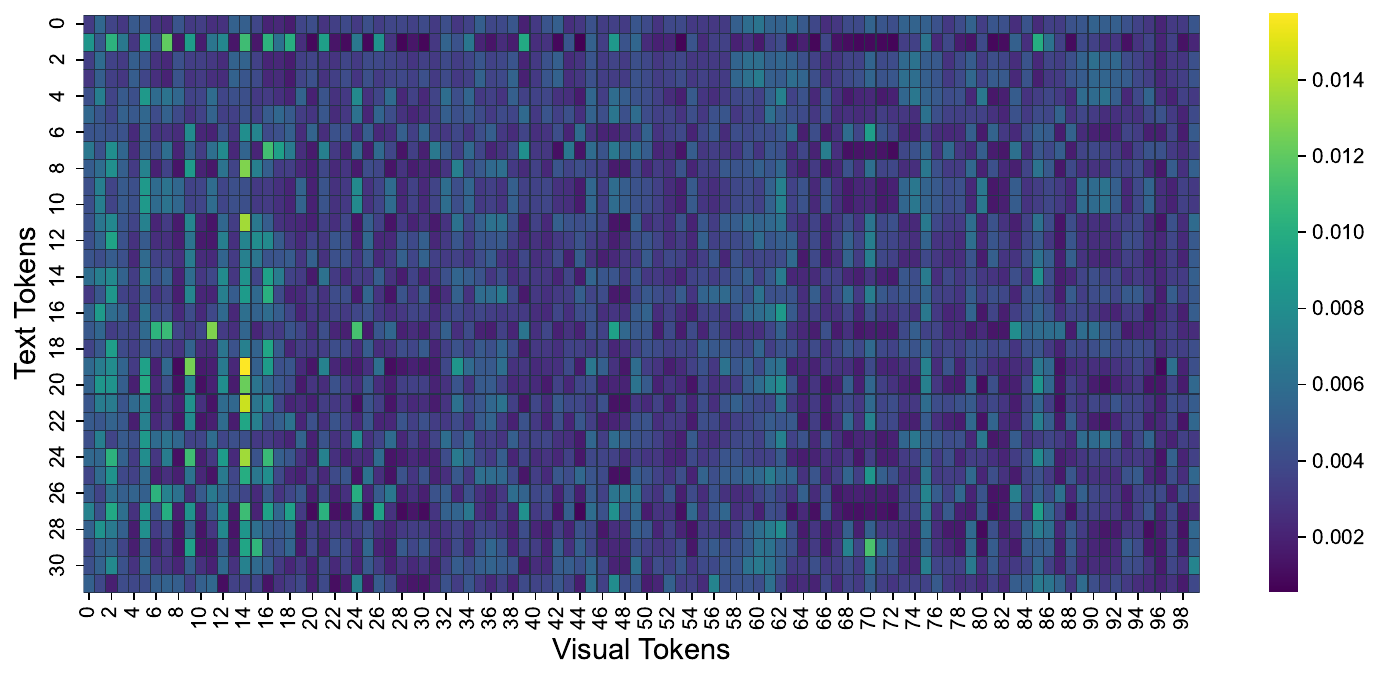} 
        \caption{}
        \label{fig:visual_attn}
    \end{subfigure}
    \begin{subfigure}{0.247\textwidth}  
        \includegraphics[width=\linewidth]{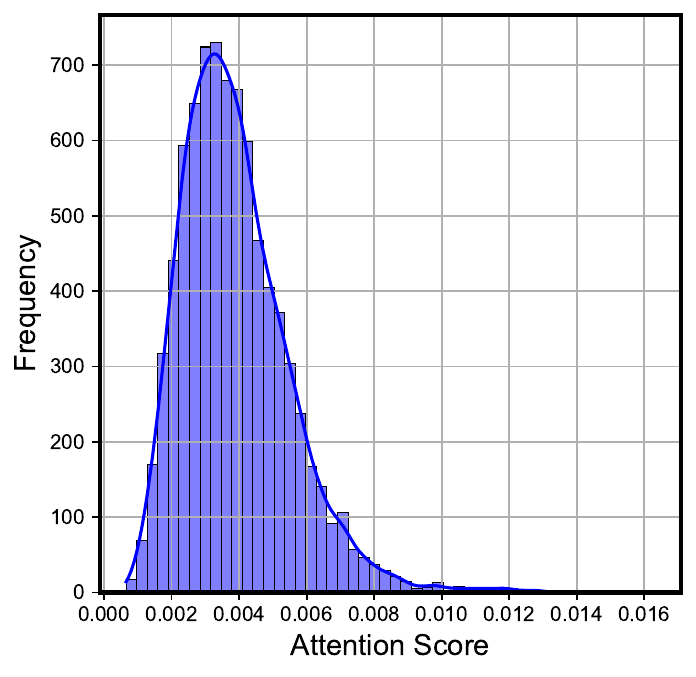} 
        \caption{}
        \label{fig:intra_frame}
    \end{subfigure}
    \begin{subfigure}{0.25\textwidth}  
        \includegraphics[width=\linewidth]{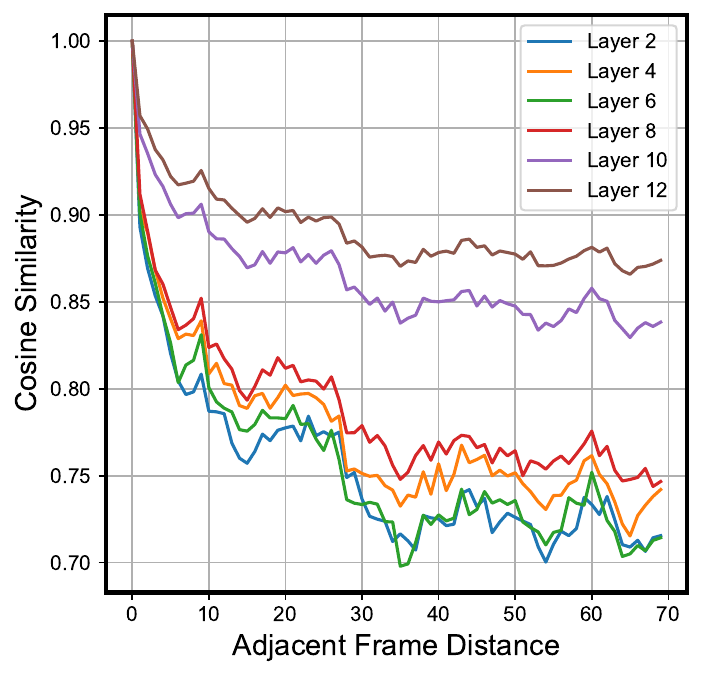} 
        \caption{}
        \label{fig:adj_frame_sim}
    \end{subfigure}
    \vspace{-3mm}
    \caption{Visualization for cross-modality attention, generated using a randomly sampled video from the MSR-VTT \cite{7780940} dataset.  (a) Cross-attention score map of the 74th frame in the final layer and last head.  (b) Cross-attention score distribution of the 80th frame in the final layer and last head. (c) The layer-wise cosine similarities of attention scores between the current frame and adjacent frames. } 
    \label{fig:example}
    \vspace{-2mm}
\end{figure*}

To solve the \textit{large memory consumption} challenge, many approaches focus on compressing video tokens. For instance, MA-LMM~\cite{he2024malmm} employs a memory bank to compress visual tokens based on the cosine similarities of adjacent two frames. Koala~\cite{TanKoala2024} passes multiple segments of video into tokenizer functions that aggregate visual tokens to handle long videos. Even though those methods reduce memory consumption, they still suffer from two significant limitations. \ul{1) \textit{Ignoring text-driven information:}} As shown in Figure \ref{fig:overview}, existing works compress visual information without considering textual information, leading to the loss of vital visual tokens that are highly related to the text prompt, particularly in Visual Question Answering (VQA)~\cite{agrawal2015visual, VQA2023, Yang_2016_CVPR, KazemiE17} tasks with complex text prompts. \ul{2) \textit{The lack of similarity adaptability:}} Previous methods attempt to compress visual tokens within the fixed and predefined frame interval, failing to capture the dynamic nature of video information and leading to inefficient and inflexible memory reduction. As shown in Figure \ref{fig:adj_frame_sim}, the similarities of adjacent frames vary across different layers. Moreover, temporally distant frames still exhibit high similarity in deep layers. Consequently, if not appropriately addressed, such limitations will severely hinder their performance in many practical applications.

To address the above limitations, in this paper, we propose \abbr, on understanding extremely long-term video with adaptive cross-modality memory reduction. As illustrated in Figure \ref{fig:overview}, our key idea is to \ul{\textit{adaptively preserve a certain number of crucial visual tokens that are most relevant to text queries across different layers based on cross-modality attention}}. We introduce two intriguing observations: \ul{1)} Only a subset of visual key states exhibit high correlations with text query; \ul{2)} Correlation varies across different layers. Our observations open opportunities for effective memory reduction by fully exploiting the inherent cross-modal capabilities of the existing visual encoders to compress visual tokens. Specifically, we first extract visual representations from video frames using a frozen visual encoder, i.e., Q-Former \cite{li2023blip2}. Then, to solve the substantial memory consumption challenge and model long-term temporal connection, we propose \abbr~attention to learn the ``query" regressively in a frame-by-frame manner within the Q-Former. \abbr~attention mechanism enables different numbers of visual tokens preserved across different layers based on cross-modality correlation. Finally, the learned queries are sent to LLM to generate the answer. To the best of our knowledge, \abbr~is \textit{the first} framework that leverages cross-modal attention integrated into the visual encoder to enable extremely long-term video understanding with efficient memory reduction. Moreover, our \abbr~can improve BLIP-based~\cite{li2022blip, li2023blip2, instructblip} models in a plug-and-play manner, enhancing their capability to process long-term video effectively. Our core contributions are summarized as follows:

\begin{itemize}
    \item We propose an efficient video memory reduction framework, \abbr~for long-term video understanding based on cross-modality attention, which, for the first time, facilitates interaction between visual features and text prompts in video understanding. 

    \item We present a cross-modality attention module that adaptively analyzes and evaluates visual tokens according to the correlation with input text prompts, enabling extremely long-term video tasks with a dynamic crucial video token-preserving strategy. 

    \item We conduct experiments on multiple video understanding tasks. Our proposed \abbr~ achieves 4.5\% accuracy improvement across multiple tasks in the LVU~\cite{lvu2021} dataset. Particularly, \abbr~ exhibits promising performance on the VQA and video captioning tasks. More importantly, \abbr~ reduces GPU memory consumption by 65\%, highlighting superior performance and efficiency in practice.
\end{itemize}

\section{Related Work}

\textbf{Video-Language Models.} With the recent advancements of large language models (LLMs)~\cite{ouyang2022instructgpt, radford2019gpt2,brown2020gpt3,touvron2023llama}, video-language models have been integrated by LLMs with image encoders for multi-modal understanding and reasoning~\cite{alayrac2022flamingo, openai2023gpt4, VideoLLM-online}. BLIP-2~\cite{li2023blip2} introduces a lightweight querying transformer~\cite{li2023blip2} to bridge the modality gap between the frozen pre-trained image encoder and LLMs. InstructBLIP~\cite{instructblip} further extracts informative visual features tailored to the given instruction with instruction-aware Query Transformer. LLaVA~\cite{damonlpsg2023videollama} leverages language-model to generate multi-modal instruction-following data and improves model generalization ability. VisionLLM~\cite{VideoLLM-online} provides a
unified perspective for vision and language tasks by treating
images as a foreign language. It aligns vision-centric
tasks with language tasks, which can be flexibly defined and
managed using language instructions. However, these models are prone to significant memory overhead when applied to long video understanding tasks.

\textbf{Long-Term Video Understanding.} Long video understanding focuses on detecting long-range patterns in videos longer than 30 seconds or even several minutes. To reduce memory and computational requirements, \cite{he2024malmm, song2024moviechat} reduce the redundancy of visual information based on the cosine similarities of adjacent visual tokens. Other works like Vis4mer~\cite{islam2022long} leverage a standard transformer encoder for short-range spatiotemporal feature extraction and a multiscale temporal Selective Structured State-Spaces (S4)~\cite{S4} decoder for long-range temporal reasoning. Koala~\cite{TanKoala2024} splits a long video into multiple segments and then aggregates visual tokens to process long videos. Considering that the goal of long-term video understanding is to answer the text question corresponding to the video, our method considers the correlation between the visual features and the input text information based on adaptive cross-modality attention, significantly reducing memory consumption and enabling extremely long-term video understanding.

\textbf{KV Cache Eviction.}
KV cache eviction, a memory reduction method that retains important context key-value pairs, is widely adopted in LLMs inference. H$_\text{2}$O~\cite{h2o_heavy_hitter_oracle} finds that keeping the recent tokens, together with ``heavy-hitter" tokens measured by attention scores, is sufficient to maintain LLM's performance. Similarly, KeyFormer~\cite{2023keyformer} retains only the key tokens in the KV cache by identifying these crucial tokens through a novel scoring function. In addition, based on the observation that KV cache states are highly similar between adjacent layers in the middle-to-deep sections of LLM, MiniCache~\cite{minicache} compresses the KV cache across layers using a novel depth-based approach, significantly reducing the memory footprint for LLM inference. Building on the success of the eviction-based method in managing long-context LLMs, we efficiently process long-term videos based on a similar philosophy. Moreover, our method focuses on reducing the redundancy of visual tokens in the long video understanding task, which is more challenging due to more information contained in videos and the interactions between visual and text modalities.

\begin{figure*}[t]  
    \centering
    \begin{subfigure}{0.98\textwidth}  
        \includegraphics[width=\linewidth]{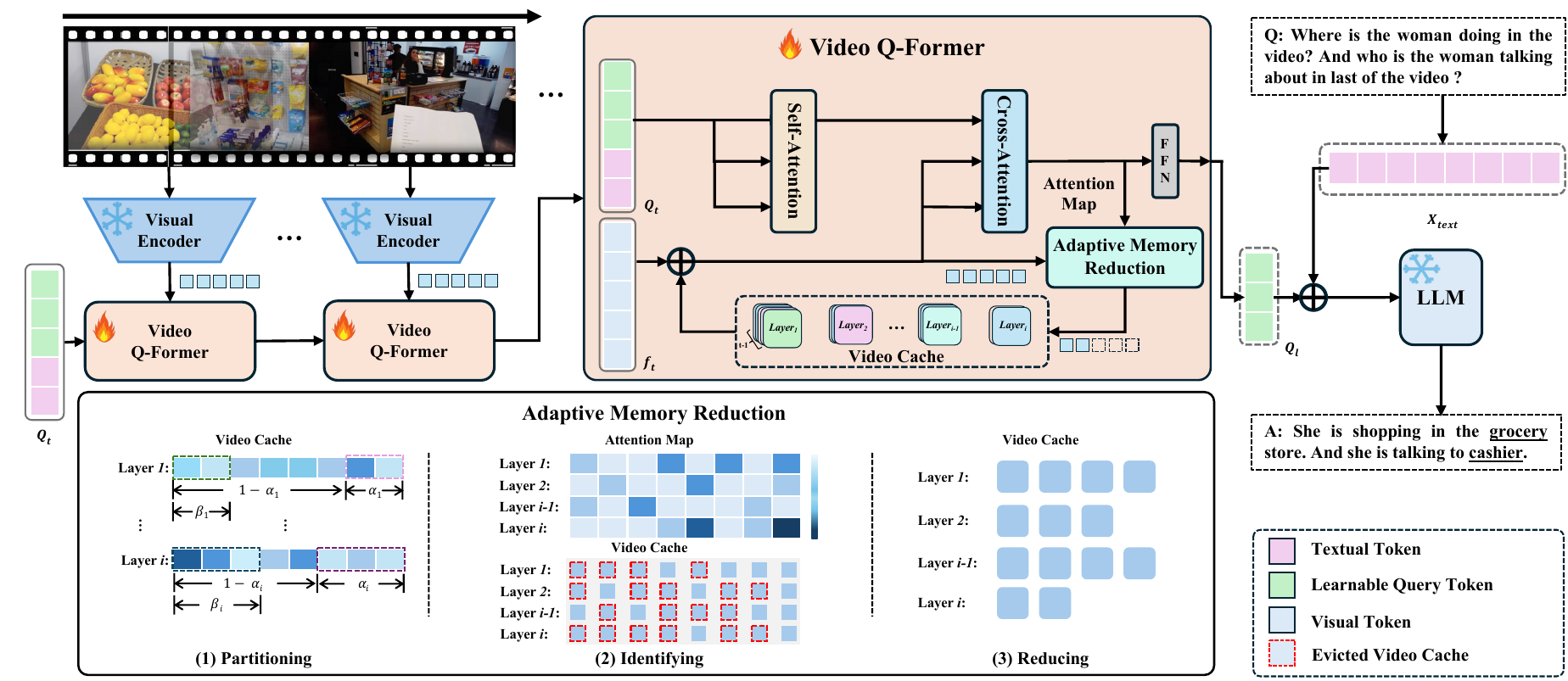} 
    \end{subfigure}
    \caption{The framework of \abbr. With video and text query as input, \abbr~first utilizes a visual encoder to extract visual features from video frames. Then, video Q-Former embeds the correlation between visual features and the text prompt into a learnable query in a regressive manner. Finally, LLM generates the answer based on the length-limited query embedding. To reduce memory consumption challenge during the process of \textit{Adaptive Memory Reduction}, the \textit{Video Cache} is partitioned into previous and recent parts. Based on cross-modality attention score, \abbr~then identifies important visual features and removes layer-wise unimportant visual tokens from cache. The snowflake denotes frozen pre-trained models, while the fire tag represents models that are fine-tuned. }
    \label{fig:arch}
\end{figure*}
\section{Observations}
\label{sec:observations}
To investigate the main bottleneck that hinders long-term video understanding, we conduct a comprehensive study on the process between video frames and text prompts, which generally perform in the Q-Former. This process aligns visual and textual information and causes substantial memory consumption when video lengths increase.
In this section, we will first analyze the cross-attention sparsity within a frame and then show the generalization of cross-attention sparsity across videos and layers. These observations demonstrate the redundancy in the dual-modality processing and inspire the foundation of our approach \abbr~proposed in Section \ref{section4}.  

\subsection{Intra-Frame Cross-Attention Sparsity}
Existing approaches compress visual tokens solely based on the similarities among video frames. However, those approaches may miss key visual tokens highly related to textual input, leading to accuracy loss for tasks with complex text questions.  Motivated by \cite{xiao2023streamingllm, ge2024model}, which have shown that retaining only a subset of salient tokens responsible for the majority of attention scores can be sufficient to maintain the performance of LLMs, we observe the similar cross-attention sparsity in the visual-textual alignment process.

\textbf{Observation 1: Only a subset of visual tokens exhibits high correlations to the text query within a frame.}\label{ob1} In Figure \ref{fig:visual_attn}, we visualize the cross-attention scores for the 74-th frame in the final layer and the last attention head by performing inference on Q-Former with randomly sampling data from the MSR-VTT \cite{7780940} dataset. The dark color indicates that the visual token exhibits a slight cross-modality correlation with the text token. We can observe that only a subset of visual tokens within a frame exhibits high correlations to text tokens. Moreover, the cross-attention scores exhibit a normal distribution as depicted in Figure \ref{fig:intra_frame}, with significant correlations primarily observed in the tail values. This observation motivates us to develop an algorithm that identifies necessary visual tokens according to cross-modality attention scores, thereby preserving crucial visual information most important to the textual prompts.

\subsection{Layer-Wise Cross-Attention Similarity}
We have demonstrated the cross-attention sparsity within a video frame. Due to the significantly increased frame length, a long-term video contains a substantial number of visual tokens. These tokens are grouped and undergo multi-layer cross-attention operations with prompt tokens. To reduce the memory consumption of redundant visual tokens globally, we further analyze cross-attention sparsity across frames and layers.

\textbf{Observation 2: Correlation varies across different layers.}\label{ob2} Based on the BLIP2 \cite{li2023blip2} Model, we conduct zero-shot inference across multiple datasets and tasks. Figure \ref{fig:adj_frame_sim} reveals that the cross-attention scores exhibit high similarity between adjacent frames, where cosine similarity exceeds 90\% among the recent five frames. More importantly, the similarity is more significant in deeper layers than in shallow layers. This observation highlights the consistent redundancy throughout the video and the diverse levels of redundancy across different layers. Correspondingly, we propose to \textit{adaptively} reduce visual memory consumption across different layers according to layer-wise cross-modality attention, making memory compression dynamic and flexible.
\section{Methodology: \abbr }\label{section4}
We present \abbr, an adaptive cross-modality memory reduction framework for extremely long-term video understanding. \abbr~consists of three stages, including \ul{1)} video feature extraction by a visual encoder; \ul{2)} adaptive memory reduction based on cross-modality attention with visual-textual embedding alignment; \ul{3)} text generation with a large language model, as illustrated in Figure \ref{fig:arch}. \abbr~adaptively reduces peak memory consumption by regressively generating learnable query tokens that preserve the temporal continuity of video and a layer-wise cross-modality memory reduction algorithm. 

\subsection{Video Feature Extraction}
\label{step:1} As shown in Figure \ref{fig:arch}, similar to common video understanding workflow~\cite{li2023blip2}, \abbr~extracts video features using a pre-trained visual encoder. Given a video with a sequence of $T$ frames, the encoder first encodes each frame and generates the corresponding video features $\bm{X} = [\bm{x}_1, \bm{x}_2, \bm{x}_3, \cdots, \bm{x}_T]$, where $\bm{x}_t \in \mathbb{R}^{P \times C}$ is the frame features at time $t$, $P$ and $C$ denote the number of visual tokens about each frame and the channel dimension of each token, respectively. Then, to incorporate temporal information into the frame-level features, a positional embedding is applied as follows: 
\begin{align}
    \bm{f}_{t} = \bm{x}_{t} + \mathcal{E}(t), \bm{f}_{t} \in \mathbb{R}^{P \times C},
\end{align}
where $\mathcal{E}(\cdot)$ represents the position embedding of a frame, and $\bm{f}_{t}$ indicates the frame features with temporal information at time $t$.

\subsection{Adaptive Memory Reduction with Cross-Modality Attention}

After extracting visual features from the video, a Q-Former is leveraged to align visual and textual features. By learning a query $\bm{Q}_l \in \mathbb{R}^{N_{l}\times C}$ in the Q-Former model, the visual features are refined to align the text description based on multi-layers cross-modal attention mechanisms, where $N_{l}$ is the number of learnable query tokens and $C$ denotes the number of feature channels.

\textbf{Visual-Textual Feature Alignment in a Regressive Manner}. Unlike existing methods that directly process all frames into the Q-Former and align visual and textual information by one shot, we propose to learn the query $\bm{Q}_l$ regressively in a frame-by-frame manner, enabling the reduction of irrelevant visual tokens based on the cross-modality correlation in a limited memory size. 

\abbr~utilizes \textit{video cache} and current visual features to align text features. To be specific, let $\bm{K}_{t}\in \mathbb{R}^{tP \times C}$, $\bm{V}_{t}\in \mathbb{R}^{tP \times C}$ to represent \textit{video cache} at time $t$, which are stored in memory by visual tokens before time $t$ and at time $t$ as: 
\begin{align}
    \bm{K}_t &= [\bm{K}_\text{t-1}, \bm{f}_t\bm{W}_{K}  ], \\
    \bm{V}_t &= [\bm{V}_\text{t-1}, \bm{f}_t\bm{W}_{V}],
\end{align}
where $\bm{W}_{K}$ and $\bm{W}_{V}$ are weight matrics. Therefore, the cross-modality attention calculation in the Q-former can be defined as
\begin{align}
    \bm{A}_{t} & = \bm{S}_{t} \cdot \bm{V}_{t} = \mathrm{softmax}\left(\frac{\bm{Q}_{t} \cdot \bm{K}_{t}^{T}}{\sqrt{C}}\right) \cdot \bm{V}_{t},
\end{align}
where $\bm{S}_{t}\in \mathbb{R}^{N \times tP}$ is the attention score, $\bm{Q}_{t}\in \mathbb{R}^{N \times C}$ is the concatenation of the learnable query $\bm{Q}_{l}$ and the text embedding $\bm{X}_\mathrm{text}$, $N$ is the number of query and text tokens.

\textbf{Layer-Wise Video Memory Reduction.} Storing all visual tokens in the \textit{video cache} is impossible due to the increasingly large memory cost while processing video frames, especially for long video understanding tasks. As shown in Observation \ref{ob1}, there is substantial visual redundancy in videos where only a subset of visual tokens exhibit high correlations with text tokens. Based on this observation, in \abbr, we propose a cross-modality attention mechanism to reduce memory consumption according to the visual-textual redundancy. \abbr~first identifies important visual features based on cross-modality attention score and then adaptively reduces video memory by removing layer-wise visual tokens not less correlated to textual information.
\begin{figure}[t]
    \centering
    \includegraphics[width=\linewidth]{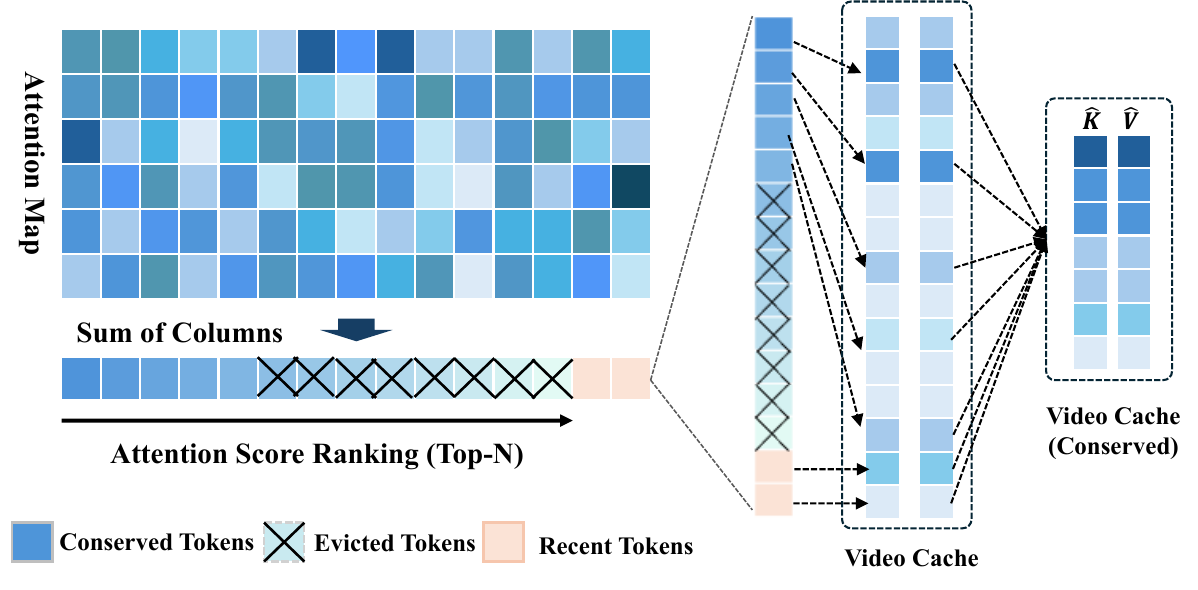}
    \vspace{-7mm}
    \caption{Illustration for our video memory reduction. The \textit{video cache} is first partitioned into recent and previous parts. Important visual tokens with high cross-modality attention scores in the previous cache are then preserved.}
    \label{fig:comp}
\end{figure}

\textit{\Circled{1} Identifying Important Visual Features based on Cross-Modality Attention Score.} 
For a video token $\bm{f}_t(i)$ at spatial location $i\in\{1,\cdots, tP\}$, the cross-modality attention score $\bm{S}^{c}_{t}(i)$ accumulates the attention scores between $\bm{f}_t(i)$ and text tokens, which can be formulated as
\begin{align}
    \bm{S}^{c}_{t}(i) = \sum_{j = 1}^{j = N} \bm{S}_{t}(j, i).
\label{cross_mdality_score}
\end{align}
Cross-modality attention scores capture the correlation between visual and textual features. Therefore, we identify a video token as pivotal if it receives a high cross-modality attention score.

\textit{\Circled{2} Adaptively Reducing Video Cache Layer-Wise.} As demonstrated in Observation \ref{ob2}, visual features exhibit different levels of similarity with textual features across different layers. Guided by cross-modal attention score, we further design an adaptive video memory reduction algorithm for the \textit{video cache}, as shown in Figure \ref{fig:comp}.

According to the order that visual token stored in the \textit{video cache}, at time $t$, we split the $\bm{K}_{t}$ for the current layer into $\textit{previous cache}$, $\hat{\bm{K}_{t}}$, and $\textit{recent cache}$, $\Tilde{\bm{K}_{t}}$:
\begin{equation}
\begin{aligned}
    &\bm{K}_t = [\hat{\bm{K}}_t, \tilde{\bm{K}}_t], \\
    &|\hat{\bm{K}}_t|/|\tilde{\bm{K}}_t| = (1 - \alpha) / \alpha 
\end{aligned}
\end{equation}
where $|\cdot|$ measures the number of tokens in cache, $\alpha$ is the split ratio. 
For the visual tokens in $\textit{recent cache}$, $\Tilde{\bm{K}_{t}}$, that reserves the latest information, we still keep it in the memory. On the other hand, for the $\textit{previous cache}$, $\hat{\bm{K}_{t}}$, we only retain visual tokens with top-$\beta$ cross-modality attention scores in memory and thus obtain the conserved cache $\overline{\bm{K}}_{t}$. The process is defined as follows:
\begin{equation}
\begin{aligned}
     &\overline{\bm{K}}_{t}=\left\{\hat{\bm{K}}_{t}(i) \mid i \in \mathrm{argtop}(\hat{\bm{S}}^{c}_{t}, \beta)\right\},\\ &|\overline{\bm{K}}_{t}|/|\hat{\bm{K}}_{t}|=\beta,
\label{cache_compression}
\end{aligned}
\end{equation}
where $\mathrm{argtop}(\cdot, \beta)$ denotes the indices of visual tokens with top-$\beta$ cross-modality attention scores, $\hat{\bm{S}}^{c}_{t}$ is the cross-modality attention scores of tokens in $\hat{\bm{K}}_{t}$, $\beta$ is the conserve ratio. Finally, the $\textit{video cache}$ $\bm{K}_{t}$ can be compressed as $\bm{K}_t = [\overline{\bm{K}}_{t}, \tilde{\bm{K}_t}]$. 
Based on our Observation \ref{ob2}, i.e., redundancy across layers varies, our adaptive video memory reduction is performed layer-wise. Correspondingly, we set $\alpha$'s and $\beta$'s for different layers. It is worth noting that the adaptive reduction is also applied to another type of $\textit{video cache}$, $\bm{V}_{t}$.

Let $r=\alpha+(1-\alpha)\beta\leq1$. Through deductive reasoning, the final size of $\textit{video cache}$ $\bm{K}_{T}$ can be derived as:
\begin{align}
    |\bm{K}_{T}|=P\sum_{t=1}^Tr^t =\frac{Pr\left(1-r^T\right)}{1-r}
\label{memory_consumption}
\end{align}
With the continuous increase of video duration, $T\rightarrow \infty$, $|\bm{K}_{T}|$ is converged to a constant $Pr/(1-r)$, showing the maximum memory requirement of \abbr~to processing unbounded video length.


\subsection{Text Generation}
With the regressive processing of video frames, the learnable query $\bm{Q}_l\in \mathbb{R}^{N \times C}$ in Q-Former has modeled the long-term temporal connection from the input video. Then, the LLM will generate the answer based on the learned $\bm{Q}_l$, a length-limited vector aggregating the long-term input video information. Specifically, assume $\bm{Y} = \{y_{1}, y_{2}, \cdots, y_{M}\}$ is the answer comprising a sequence of $M$ words, we minimize the cross-entropy loss during training as follows:
\begin{align}
    L(\bm{V}, \bm{P}, \bm{Y}) = -\sum_{i=1}^{M}y_i\log p(y_{i}|y_{<i}, \bm{Q}_l, \bm{X}_\mathrm{text})
\end{align}
where $p(y_{i}|y_{<i}, \bm{Q}_l, \bm{X}_\mathrm{text})$ denotes the probability for the $i$-th word $y_{i}$ given the preceding sequence of ground truth words $y_{<i}$. 

\begin{table*}[!ht]
\centering
\begin{minipage}[t]{0.95\linewidth}
    \centering
    \caption{Comparison with state-of-the-art methods on the LVU dataset. The \underline{underlined} number means the second best.}
    \vspace{-3mm}
    \label{table: lvu}
\begin{tabular}{l|>{\centering\arraybackslash}p{1.2cm}>{\centering\arraybackslash}p{1.2cm}>{\centering\arraybackslash}p{1.2cm}|>{\centering\arraybackslash}p{1.2cm}>{\centering\arraybackslash}p{1.2cm}>{\centering\arraybackslash}p{1.2cm}>{\centering\arraybackslash}p{1.2cm}|c}
    \toprule
    \multicolumn{1}{c|}{\multirow{2}[4]{*}{\textbf{Model}}} & \multicolumn{3}{c|}{{\textbf{Content}}} & \multicolumn{4}{c|}{{\textbf{Metadata}}} & \multicolumn{1}{c}{\multirow{2}[4]{*}{\textbf{Avg}}} \\
    \cmidrule{2-8}          
    & Relation & Speak & Scene & Director & Genre & Writer & \multicolumn{1}{c|}{Year} &  \\
    \midrule
    {Obj\_T4mer~\cite{lvu2021}} & 54.8  & 33.2  & 52.9  & 47.7  & 52.7  & 36.3  & \multicolumn{1}{c|}{37.8} & 45.1 \\
    {Performer~\cite{Performer}} & 50.0  & 38.8  & 60.5  & 58.9  & 49.5  & 48.2  & \multicolumn{1}{c|}{41.3} & 49.6 \\
    {Orthoformer~\cite{patrick2021keeping}} & 50.0  & 38.3  & 66.3  & 55.1  & 55.8  & 47.0  & \multicolumn{1}{c|}{43.4} & 50.8 \\
    {VideoBERT~\cite{VideoBERT}} & 52.8  & 37.9  & 54.9  & 47.3  & 51.9  & 38.5  & \multicolumn{1}{c|}{36.1} & 45.6 \\
    {LST~\cite{islam2022long}} & 52.5  & 37.3  & 62.8  & 56.1  & 52.7  & 42.3  & \multicolumn{1}{c|}{39.2} & 49.0 \\
    {VIS4mer~\cite{islam2022long}} & 57.1  & 40.8  & 67.4  & 62.6  & 54.7  & 48.8  & \multicolumn{1}{c|}{44.8} & 53.7 \\
    {S5~\cite{wang2023selective}} & \textbf{67.1} & \underline{42.1}  & 73.5  & 67.3  & \underline{65.4}  & 51.3  & \multicolumn{1}{c|}{48.0} & 59.2 \\
    {MA-LMM~\cite{he2024malmm}} & 58.2  & \textbf{44.8} & \underline{80.3}  & \underline{74.6}  & 61.0  & \underline{70.4}  & \multicolumn{1}{c|}{\underline{51.9}} & \underline{63.0} \\
    \midrule
    {\textbf{Ours}} & \underline{63.1}  & 40.2 & \textbf{86.2} & \textbf{75.4} & \textbf{68.0} & \textbf{77.0} & \textbf{62.5} & \textbf{67.5} \\
    \midrule
    \cmidrule{2-9}    
\end{tabular}

    \vspace{-2mm}
\end{minipage}

\vspace{4pt} 

\begin{minipage}[t]{0.48\linewidth}
    \centering
    \caption{Comparison on the Breakfast and COIN datasets. Top-1 accuracy is reported.}
    \vspace{-3mm}
    \label{tab:vqa_long}
    \renewcommand{\arraystretch}{1.14}
\begin{tabular}{l|>{\centering\arraybackslash}p{2cm}>{\centering\arraybackslash}p{2cm}}
    \toprule
    \textbf{Model} & \textbf{Breakfast} & \textbf{COIN} \\
    \midrule
    TSN~\cite{TSN}   & -   & 73.4 \\
    VideoGraph~\cite{videograph} & 69.5  & - \\
    Timeception~\cite{Timeception} & 71.3  & - \\
    GHRM~\cite{GHRM}  & 75.5  & - \\
    D-Sprv.~\cite{D-Sprv} & 89.9  & 90.0 \\
    ViS4mer~\cite{islam2022long} & 88.2  & 88.4 \\
    S5~\cite{wang2023selective}    & 90.7  & 90.8 \\
    MA-LMM~\cite{he2024malmm} & \underline{93.0}  & \underline{93.2} \\
    \midrule
    {\textbf{Ours}} & \textbf{94.4} & \textbf{93.3} \\
    \bottomrule
\end{tabular}

\end{minipage}
\hfill
\begin{minipage}[t]{0.48\linewidth}
    \centering
    
    \caption{Comparison on the MSRVTT, MSVD and YouCook2 datasets.}
    \vspace{-3mm}
    \label{table: captioning}
    \renewcommand{\arraystretch}{1.1}
\begin{tabular}{>{\raggedright\arraybackslash}p{2.37cm}|>{\centering\arraybackslash}p{0.3cm}>{\centering\arraybackslash}p{0.5cm}|>{\centering\arraybackslash}p{0.3cm}>{\centering\arraybackslash}p{0.7cm}|>{\centering\arraybackslash}p{0.3cm}>{\centering\arraybackslash}p{0.5cm}}
    \toprule
    \multicolumn{1}{c|}{\multirow{2}[4]{*}{\textbf{Model}}} & \multicolumn{2}{c|}{\makecell{\textbf{MSRVTT}}} & \multicolumn{2}{c|}{\makecell{\textbf{MSVD}}} & \multicolumn{2}{c}{\textbf{YouCook2}} \\
    \cmidrule{2-7}
    \multicolumn{1}{c|}{} & \multicolumn{1}{c}{\textbf{M}} & \textbf{C} & \multicolumn{1}{c}{\textbf{M}} & \textbf{C} & \multicolumn{1}{c}{\textbf{M}} & \textbf{C} \\
    \midrule
    {UniVL~\cite{UniVL}} &28.2&49.9&29.3&52.8&-&127.0\\
    {SwinBERT~\cite{SwinBERT}} & 29.9  & 53.8  & 41.3  & 120.6 & 15.6  & 109.0\\
    {GIT~\cite{GIT}} & 32.9  & 73.9  & \underline{51.1} & \underline{180.2} & \underline{17.3}  & \underline{129.8}\\
    {mPLUG-2~\cite{mPLUG-2}} & \textbf{34.9} & \textbf{80.3} & 48.4  & 165.8 & -   & -\\
    {VideoCoca~\cite{VideoCoCa}} & -   & 73.2  & -   & -   & -   & 128.0\\
    {VideoLLaMA~\cite{Video-LLaMA}}& 32.9  & 71.6  & 49.8  & 175.3 & 16.5  & 123.7\\
    {MA-LMM~\cite{he2024malmm}} & \underline{33.4}  & \underline{74.6}  & 51.0    & 179.1 & \textbf{17.6} & \textbf{131.2}\\
    \midrule
    {\textbf{Ours}} & {33.0} & {73.1} &   \textbf{51.4}   &  \textbf{189.4}    & \textbf{17.6}  & 125.6\\
    \bottomrule
\end{tabular}
 
\end{minipage}
\vspace{-2mm}
\end{table*}
 \label{table: performance}
\begin{algorithm}[t]
    \caption{Layer-wise Adaptive Video Reduction.}
    \label{algo:kv-cache}
    
\begin{algorithmic}[1]
    \Input{ Video frames $\left\{\bm{f}_{t}\right\}_{t=1}^T$, hyper-parameters $\alpha, \beta$ for each layer;}
    \Output{ Reduced video KV cache $\bm{K}_{t}, \bm{V}_{t}$.}

    \For{$t = 1$ to $T$}
        \State $\bm{K}_t \leftarrow [\bm{K}_{t-1}, \bm{f}_t \bm{W}_{K} ];  \bm{V}_t \leftarrow [\bm{V}_{t-1}, \bm{f}_t\bm{W}_{V}]$;
        \State \textbf{Partitioning:} $\bm{K}_t \rightarrow [\hat{\bm{K}_{t}}, \Tilde{\bm{K}_{t}}]; \bm{V}_t \rightarrow [\hat{\bm{V}_{t}}, \Tilde{\bm{V}_{t}}];$
        \State \textbf{Identifying:} Determine $\bm{S}^{c}_{t}$ using Eq.\ref{cross_mdality_score};
        \State \textbf{Reducing:} Obtain $\overline{\bm{K}}_{t}$ using Eq.\ref{cache_compression}; 
        \State $\bm{K}_t \leftarrow [\overline{\bm{K}}_{t}, \tilde{\bm{K}_t}]; \bm{V}_t \leftarrow [\overline{\bm{V}}_{t}, \tilde{\bm{V}_t}]$;
    \EndFor
\end{algorithmic}
\end{algorithm}

\section{Experiments}

\subsection{Tasks and Datasets}

\textbf{Long-Term Video Understanding.} We conduct experiments on three common long-term video understanding datasets, including LVU \cite{lvu2021}, Breakfast \cite{breakfast} and COIN\cite{COIN}. We report the Top-1 accuracy as the evaluation metric. In particular, for the LVU dataset, we focus our experiments on seven classification tasks: relationship, speaking style, scene, director, genre, writer, and release year. These tasks provide a diverse range of challenges for evaluating model performance in different aspects of video understanding.

\textbf{Video Question Answering.} We evaluate \abbr on two popular video question answering datasets: MSRVTT-QA~\cite{7780940}, MSVD-QA~\cite{MSVD}. MSRVTT-QA and MSVD-QA contain shorter videos, ranging from 10 to 15 seconds. Regarding the dataset scale, MSRVTT-QA includes 10,002 videos, MSVD-QA has 1,971 videos. 

\textbf{Video Captioning.} We present the video captioning results with the METEOR~\cite{banerjee-lavie-2005-meteor} and CIDEr~\cite{CIDEr} metrics on three popular datasets, MSRVTT~\cite{7780940}, MSVD~\cite{MSVD}, and YouCook2~\cite{ZhXuCoAAAI18}. It is noted that YouCook2 encompasses videos longer than five minutes, posing a significant memory consumption challenge to understand the videos across such long periods.
\subsection{Implementation Details}

For the visual encoder, we utilize the pre-trained image encoder ViT-G/14~\cite{Scalling-Vit} from EVA-CLIP\cite{eva_clip_2023}. To efficiently align the visual and text features, we employ the pre-trained Q-Former weights from InstructBLIP~\cite{instructblip}.  For the LLM decoding, we use the pre-trained LLM, Vicuna-7B~\cite{vicuna2023} V1.1.
During training, we keep the visual encoder and the LLM decoder frozen, and fine-tune the trainable parameters of Q-former. For video input, we extract frames at a speed of 10 fps.
All experiments are tested on our server with eight NVIDIA RTX 6000 Ada GPUs, two AMD EPYC 9254 CPUs, and 1.5 TB memory. 

\subsection{Main Results}
\textbf{Long-Term Video Understanding.} We compare \abbr~with state-of-the-art methods on the LVU~\cite{lvu2021} benchmark. As shown in Table \ref{table: lvu}, our model outperforms the existing long-term video understanding models, including Object transformer~\cite{lvu2021}, S4~\cite{wang2023selective}, VIS4mer~\cite{islam2022long}, VideoBERT~\cite{VideoBERT}, LST~\cite{islam2022long}, and MA-LMM~\cite{he2024malmm} across both content understanding and metadata prediction tasks. The results indicate that \abbr~achieves a significant improvement across most tasks, increasing Top-1 accuracy by 4\% compared to MA-LMM~\cite{he2024malmm}.

In Table \ref{tab:vqa_long}, we evaluate our \abbr~on the Breakfast \cite{breakfast} and COIN \cite{COIN} datasets. It is worth noting that these datasets present a greater memory consumption challenge due to the longer and more diverse videos they contain. It is seen that our \abbr~outperforms MA-LMM \cite{he2024malmm} by 1.4\% and 0.1\% in Top-1 accuracy, respectively. These results demonstrate that our approach achieves superior performance in long-term video understanding.

\textbf{Video Captioning.} Table \ref{table: captioning} summarizes the experimental results on video captioning datasets, including MSRVTT, MSVD, and YouCook2. \abbr~achieves 51.4\% Meteor and 189.4\% CIDEr on the MSVD datasets. The results show that \abbr~outperforms the prior state-of-the-art approach, MA-LMM ~\cite{he2024malmm}, with gains of 0.4\% and 10.3\%, respectively. Although mPLUG-2 has slightly better performance on MSRVTT, it demands extensive data for pre-training, leading to a training overhead.

\begin{figure}[!t]
    \centering
    \includegraphics[width=0.95\linewidth]{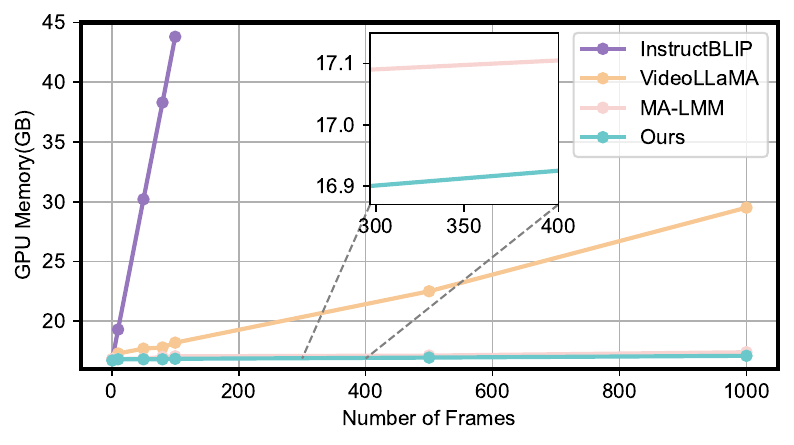}
    \vspace{-4mm}
    \caption{Practical memory consumption analysis compared to existing methods. The GPU memory usage for InstructBLIP shows an exponential increase, leading to out-of-memory (\textbf{OOM}) errors after processing \ul{100} frames. In contrast, VideoLLaMA exhibits a linear increase in memory usage. \abbr~maintains almost \ul{constant} memory usage without sacrificing performance even as frames significantly increase. }
    \label{fig:memory}
\end{figure}

\textbf{Memory Analysis.} In addition to performance evaluation, we also conduct experiments on memory usage with randomly selected videos from the LVU~\cite{lvu2021}, MSRVTT~\cite{7780940}, and MSVD~\cite{MSVD} datasets. As shown in Figure \ref{fig:memory}, existing methods like InstructBLIP~\cite{instructblip} and VideoLLaMA~\cite{Video-LLaMA} instantly exhibit rapid increases in memory consumption as the number of frames increases, leading to out-of-memory (\textbf{OOM}) errors. In contrast, \ul{\abbr~achieves a significant reduction in memory usage by nearly 65\%, and maintains almost constant memory consumption without sacrificing performance.} Furthermore, most of the occupied memory is consumed by LLM, with only a small fraction allocated for adaptive cross-modality alignment, which can be further alleviated if we use a lightweight LLM.

\subsection{Ablation Studies}

\begin{figure}[t!]
    \centering
    \includegraphics[width=\linewidth]{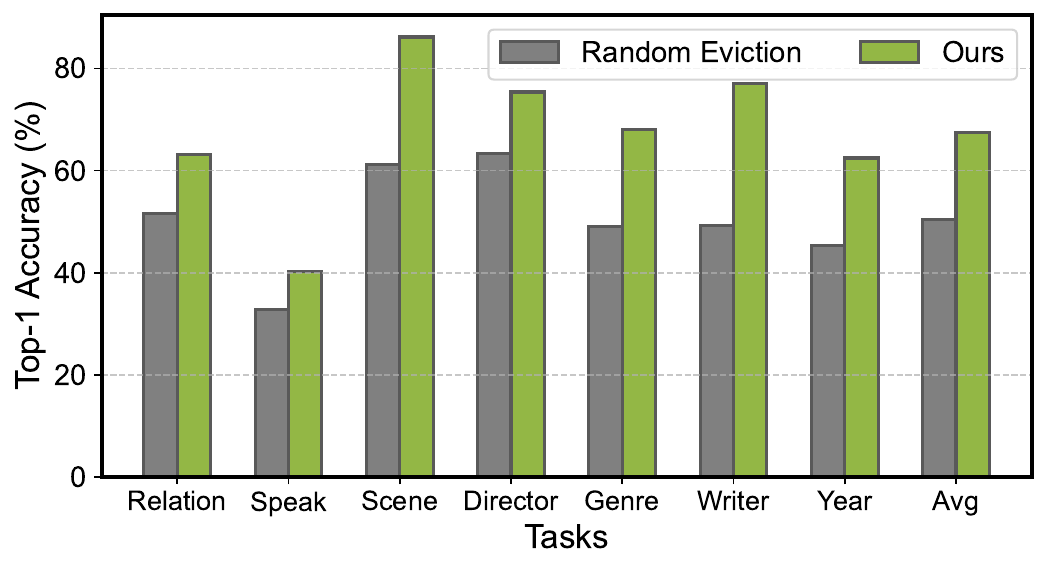}
    \vspace{-7mm}
    \caption{Performance Comparison between two memory reduction strategies, random eviction v.s \abbr. Top-1 accuracy is reported here.}
    \label{fig:ablation-1}
\end{figure}

\textbf{Memory Reduction.} Our method reduces video memory adaptively based on cross-modality attention scores, achieving extremely long-term video understanding with low memory cost. To investigate the importance of our memory reduction, we compare our cache strategy with the random eviction on the LVU dataset, which randomly discards the same number of visual tokens, as shown in Figure \ref{fig:ablation-1}. It is seen that our approach significantly outperforms the random one, indicating the effectiveness of our cross-attention-based memory reduction.

\begin{figure}[!t]
    \centering
    \begin{subfigure}{0.48\linewidth}
        \centering
        \includegraphics[width=\linewidth]{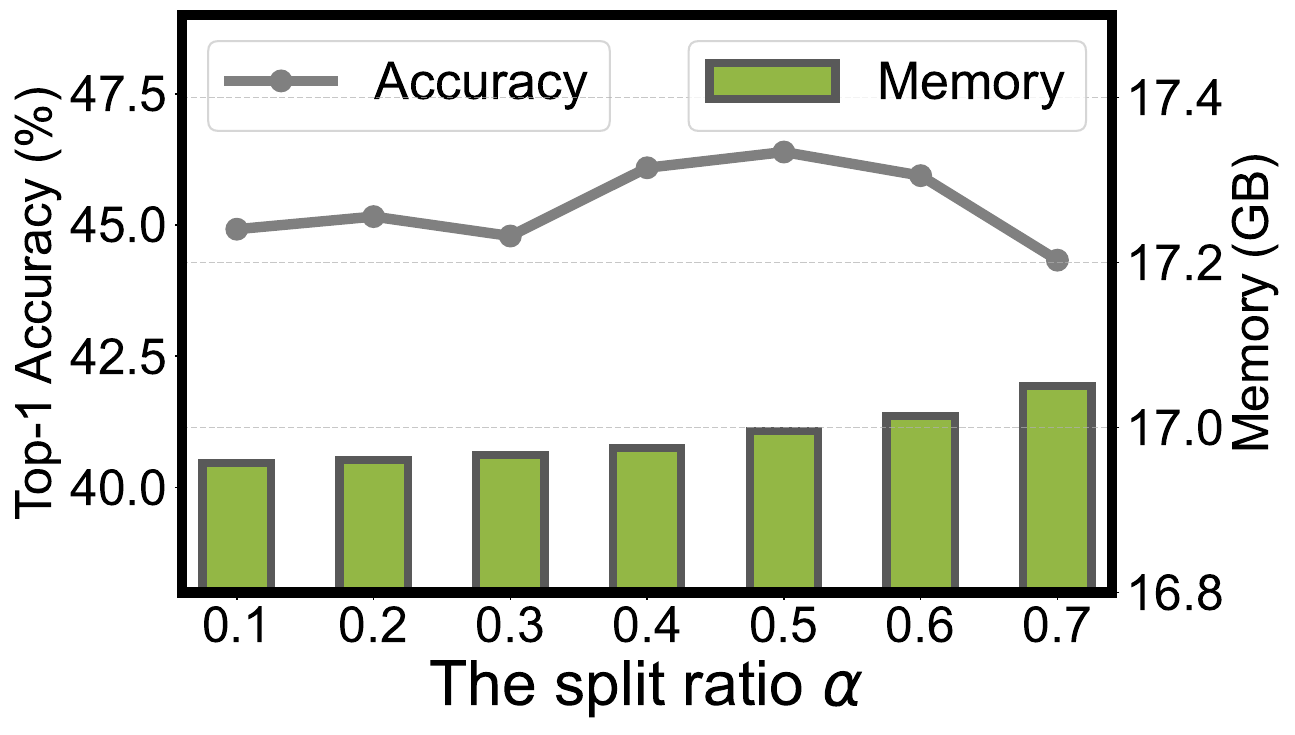}
        \caption{MSRVTT-QA dataset, $\beta$ = 0.1 }
        \label{fig:subfig1}
    \end{subfigure}
    \hfill
    \begin{subfigure}{0.48\linewidth}
        \centering
        \includegraphics[width=\linewidth]{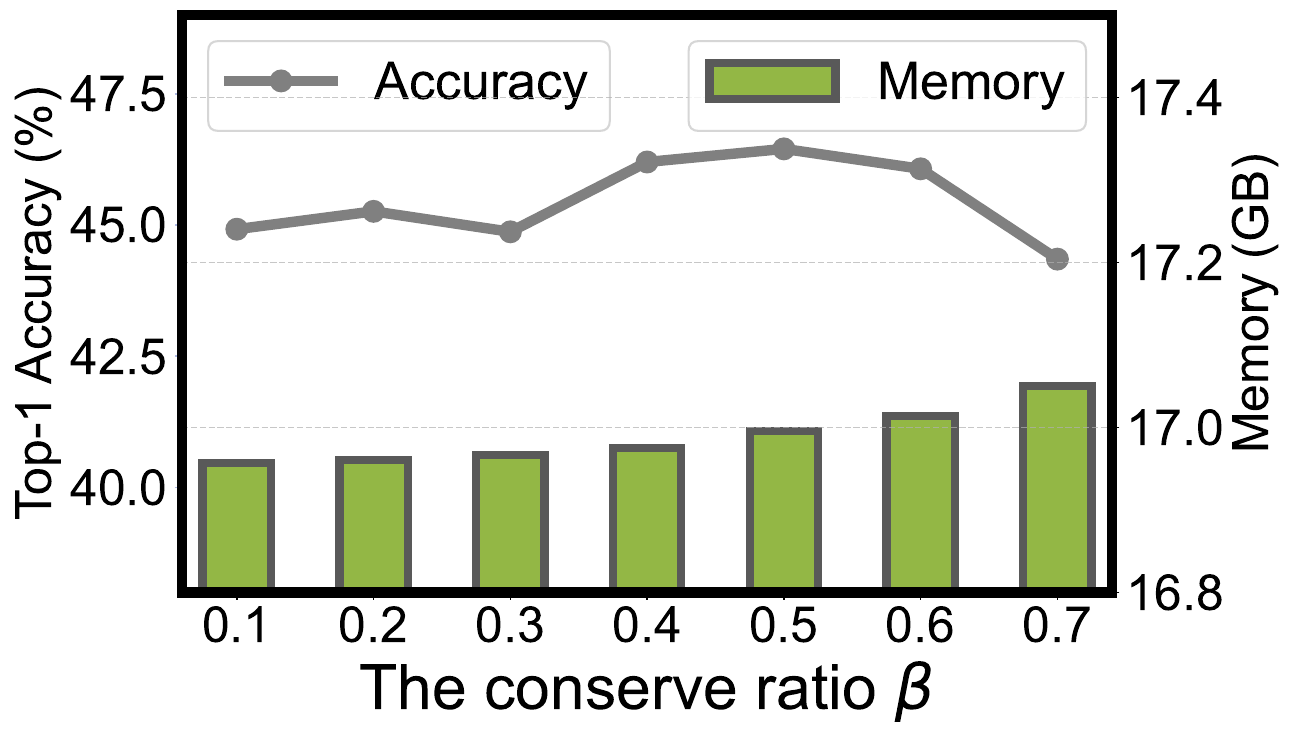}
        \caption{MSRVTT-QA dataset, $\alpha$ = 0.1}
        \label{fig:subfig2}
    \end{subfigure}

    \vskip\baselineskip 

    \begin{subfigure}{0.48\linewidth}
        \centering
        \includegraphics[width=\linewidth]{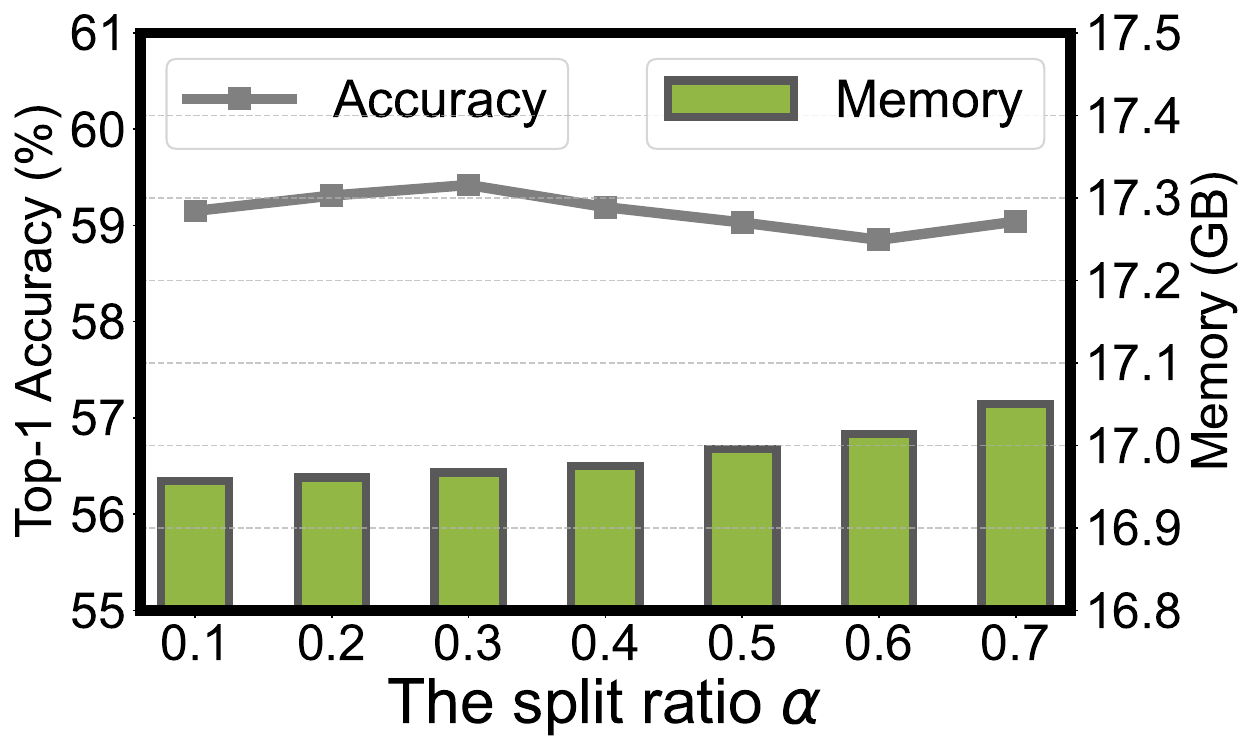}
        \caption{MSVD-QA dataset, $\beta$ = 0.1 }
        \label{fig:subfig3}
    \end{subfigure}
    \hfill
    \begin{subfigure}{0.48\linewidth}
        \centering
        \includegraphics[width=\linewidth]{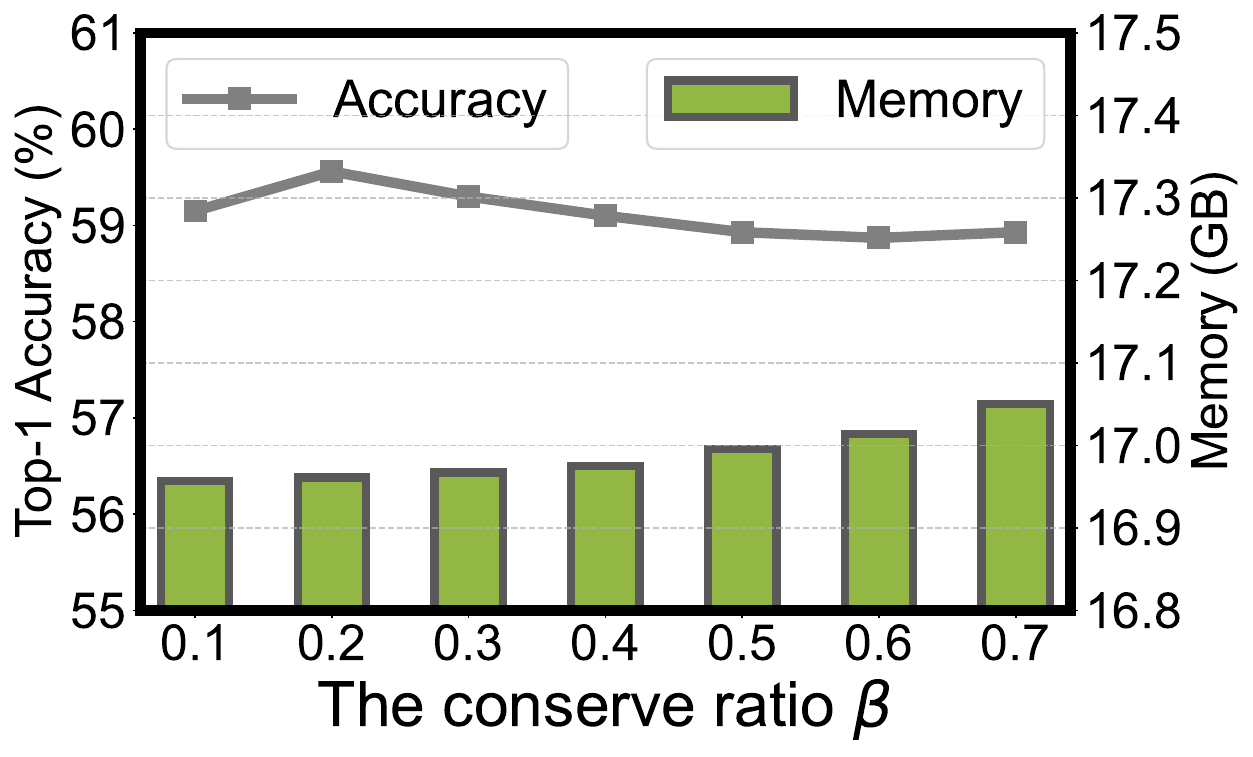}
        \caption{MSVD-QA dataset, $\alpha$ = 0.1}
        \label{fig:subfig4}
    \end{subfigure}
    \vspace{-2mm}
    \caption{Influence of split ratio $\alpha$ and conserve ratio $\beta$ on performance and memory usage.}
    \label{tab:ablation-2}
\end{figure}

\textbf{Hyperparameters Analysis.}
Split ratio $\alpha$ and conserve ratio $\beta$ are hyper-parameters controlling the preservation of critical visual tokens. In order to determine the appropriate settings, we explore the influence of $\alpha$ and $\beta$ on memory reduction and performance on MSRVTT and MSVT datasets. Figure \ref{tab:ablation-2} shows that more retaining tokens result in higher memory usage. Moreover, with the increase of $\alpha$ ($\beta$), the accuracy rises to a peak and then decreases. This decrease is due to the negative impact of retaining redundant information, which diverts the model's attention away from essential tokens. To achieve an optimal balance between performance and memory efficiency, we set $\alpha=0.1$ and $\beta=0.1$ for all layers. 

\textbf{LLM Decoding.} LLM plays an essential role in producing instruction cues. To investigate the influence of LLMs on our {\abbr}, we compare the results using multiple LLMs, including \textit{FlanT5-XL}~\cite{FlanT5-XL} and \textit{Vicuna-7B}~\cite{vicuna2023}. Table \ref{tab:ablation-3} indicates that Vicuna-7B achieves better performance in all tasks. Therefore, we choose the Vicuna-7B model as our LLM backbone. Since our \abbr~framework is designed to accommodate various modern LLMs, we plan to conduct further tests in future work.

\begin{table}[!t]
    \centering
    
    \caption{Ablation study of LLM decoding methods, with results reported for the METEOR and CIDEr metrics.}
    \vspace{-3mm}
\begin{tabular}{>{\raggedright\arraybackslash}p{1.65cm}|>{\centering\arraybackslash}p{0.55cm}>{\centering\arraybackslash}p{0.6cm}|>{\centering\arraybackslash}p{0.55cm}>{\centering\arraybackslash}p{0.8cm}|>{\centering\arraybackslash}p{0.55cm}>{\centering\arraybackslash}p{0.65cm}}
    \toprule
    \multicolumn{1}{c|}{\multirow{2}[4]{*}{\textbf{Model}}} & \multicolumn{2}{c|}{\textbf{MSRVTT}} & \multicolumn{2}{c|}{\textbf{MSVD}} & \multicolumn{2}{c}{\textbf{YouCook2}} \\
    \cmidrule{2-7}
    \multicolumn{1}{c|}{} & \multicolumn{1}{c}{\textbf{M}} & \textbf{C} & \multicolumn{1}{c}{\textbf{M}} & \textbf{C} & \multicolumn{1}{c}{\textbf{M}} & \textbf{C} \\
    \midrule
    {FlanT5-XL} &21.6$\downarrow$&58.9$\downarrow$&48.7$\downarrow$&166.9$\downarrow$&15.3$\downarrow$&107.6$\downarrow$\\
    \midrule
    {{Vicuna-7B}} & {33.0} & {73.1} &   {51.4}   &  {189.4}    & {17.6}  & 125.6\\
    \bottomrule
\end{tabular} 
    \label{tab:ablation-3}
\end{table}
\section{Conclusion}

In this paper, we present {\abbr}, an adaptive cross-modality memory reduction framework for extremely long-term video understanding. The key idea of \abbr~is to adaptively preserve a certain number of crucial visual tokens most relevant to text queries across different layers based on cross-modality attention, addressing the substantial memory consumption and modeling long-term temporal connection.
Moreover, our \abbr~enables BLIP-based~\cite{li2022blip, li2023blip2, instructblip} models in a plug-and-play manner, enhancing their capability to process long-term video effectively.
Experiments on video question understanding and captioning tasks demonstrate the superiority of the proposed \abbr~over existing state-of-the-art approaches.

{
    \small
    \bibliographystyle{ieeenat_fullname}
    \bibliography{main}
}

\end{document}